\documentclass[a4paper,10pt]{article}

\usepackage[margin=0mm]{geometry} 
\usepackage{times}
\usepackage{authblk}
\usepackage{graphicx}
\usepackage{mathtools}
\usepackage{titlesec}
\usepackage{setspace}
\usepackage{placeins}
\usepackage{float}
\usepackage{subcaption}
\usepackage{tikz}
\usetikzlibrary{arrows.meta, positioning, shapes.misc, shapes.geometric}
\usepackage{graphicx}
\usepackage{amsmath}
\usepackage{hyperref}
\usepackage{mathptmx}
\usepackage{amssymb}
\usepackage{fancyhdr}
\usepackage{mathptmx} 

\AtBeginDocument{
  \abovedisplayskip=6pt
  \belowdisplayskip=6pt
  \abovedisplayshortskip=3pt
  \belowdisplayshortskip=3pt
}

\usepackage{etoolbox}

\patchcmd{\section}{\ignorespaces}{\@afterindentfalse\ignorespaces}{}{}
\patchcmd{\subsection}{\ignorespaces}{\@afterindentfalse\ignorespaces}{}{}
\patchcmd{\subsubsection}{\ignorespaces}{\@afterindentfalse\ignorespaces}{}{}



\titleformat{\section}
  {\bfseries\fontsize{12pt}{14pt}\selectfont}
  {\thesection}{0.7em}{}

\titleformat{\subsection}
  {\bfseries\fontsize{10pt}{12pt}\selectfont}
  {\thesubsection}{0.7em}{}

\titleformat{\subsubsection}
  {\itshape\fontsize{10pt}{12pt}\selectfont}
  {\thesubsubsection}{0.7em}{}

\titlespacing*{\section}{0pt}{6mm}{3mm}
\titlespacing*{\subsection}{0pt}{6mm}{3mm}
\titlespacing*{\subsubsection}{0pt}{6mm}{3mm}

\makeatletter
\renewcommand{\maketitle}{
    \begin{center}
        {\bfseries\fontfamily{phv}\selectfont\fontsize{16pt}{18pt}\@title\par}
        \vspace{6mm}
        {\normalsize \@author \par}
    \end{center}
}
\makeatother

\title{\fontfamily{phv}\selectfont\bfseries Model-less feedback control of space-based continuum manipulators using backbone tension optimization}

\author{
    \textit{Shrreya} Rajneesh\textsuperscript{1},
    \textit{Rakesh Kumar} Sahoo\textsuperscript{2},
    and \textit{Manoranjan} Sinha\textsuperscript{3} \\
    \textsuperscript{1}Department of Aerospace Engineering, Indian Institute of Technology Kharagpur, West Bengal, 721302, India \\
    \textsuperscript{2}Department of Aerospace Engineering, Indian Institute of Technology Kharagpur, West Bengal, 721302, India \\
    \textsuperscript{3}Department of Aerospace Engineering, Indian Institute of Technology Kharagpur, West Bengal, 721302, India
}

\newenvironment{customabstract}{
\vspace*{8mm}
  \begin{list}{}{
      \setlength{\leftmargin}{-3mm}
      \setlength{\rightmargin}{-3mm}
  }
  \item[]\fontsize{9pt}{11pt}\selectfont  
}{%
  \end{list}
  \vspace*{10mm}
}

\begin{document}
\maketitle
\begingroup
\renewcommand\thefootnote{\fnsymbol{footnote}}%
\footnotetext[1]{Corresponding author: \href{mailto:shrreyarajneesh07@gmail.com}{shrreyarajneesh07@gmail.com}}
\endgroup

\begin{customabstract} 
\textbf{\fontfamily{phv}\selectfont\bfseries Abstract.}  Continuum manipulators offer intrinsic dexterity and safe geometric compliance for navigation within confined and obstacle-rich environments. However, their infinite-dimensional backbone deformation, unmodeled internal friction, and configuration-dependent stiffness fundamentally limit the reliability of model-based kinematic formulations, resulting in inaccurate Jacobian predictions, artificial singularities, and unstable actuation behavior. Motivated by these limitations, this work presents a complete model-less control framework that bypasses kinematic modeling by using an empirically initialized Jacobian refined online through differential convex updates. Tip motion is generated via a real-time quadratic program that computes actuator increments while enforcing tendon slack avoidance and geometric limits. A backbone-tension optimization term is introduced in this paper to regulate axial loading and suppress co-activation compression. The framework is validated across circular, pentagonal, and square trajectories, demonstrating smooth convergence, stable tension evolution, and sub-millimeter steady-state accuracy without any model calibration or parameter identification. These results establish the proposed controller as a scalable alternative to model-dependent continuum manipulation in a constrained environment. 
\end{customabstract}

\section{Introduction}

In recent times, continuum manipulators have emerged as a compelling alternative to conventional rigid-link robotic manipulators due to their dexterity, inherent compliance and ability to navigate constrained environments\cite{Jones2006Multisection}. The continuous backbone structure of continuum manipulators enables smooth, flexible, unhindered motion, making them fit for minimally invasive surgery\cite{Simaan2008Force}, search-and-rescue and increasingly, for human robot interaction. Lately, continuum robots have attracted significant interest in extraterrestrial inspection and servicing tasks where lightweight structures, inherent compliance and ability to operate in unstructured conditions are necessary requirements. For example, Frazelle \textit{et al.} showed how continuum manipulators can be deployed in microgravity using only external vision sensing\cite{ContinuumKMF_Frazelle}, highlighting their potential utility aboard the International Space Station for maintenance, repair, and structural inspection. Together, these developments position continuum manipulators as a leading contender for deployment in hazardous, constrained and geometrically complex domains.

Robotic manipulators as a whole have also seen major advancement in precision, control and reliability, further motivating the exploration of continuum robots\cite{Uppal2025EAPF}. A broad spectrum of continuum robot and manipulator architectures have been explored in literature, including tendon driven catheters \cite{Camarillo2008Mechanics}, multisection spatial manipulators \cite{Jones2006Multisection}, and even soft pneumatic/elastomeric robots that are actuated by fluidic chambers or structural deformation \cite{FEM2017}. The kinematic and dynamic strategies also span several domains. Classical approaches rely on model-based constant curvature models \cite{Webster2010CCReview} which offer closed form kinematics at the expense of modeling accuracy in complex environments. More mechanics based tendon beam models incorporate axial compression, stiffness coupling, and tendon geometry to improve prediction fidelity\cite{Camarillo2008Mechanics}. Other strategies also include fuzzy-model based inverse kinematic control \cite{FuzzyControl2016}, and even high fidelity finite-element based formulations \cite{FEM2017}. Studies on closed-loop control compare joint space and task space strategies \cite{ClosedLoopEval}, while advanced nonlinear controllers including PID, fuzzy logic and Sliding mode control(SMC) have been proposed for multi-section systems\cite{HRMDynamics}. Even though these methods achieve exceptional accuracy under well defined environmental conditions, many rely on precise knowledge of the geometric, mechanical, and stiffness parameters, making them highly sensitive to modeling errors and external disturbances, and thus challenging to maintain outside laboratory settings.

Despite years of progress, accurate forward and differential kinematics for continuum robots remains quite difficult to obtain in practice as contact, friction, unmodeled elastic effects, etc lead to degraded tracking performance, artificial singularities and even signs of Jacobian inversion\cite{Yip2014ModelLess, Yip2016ModelLess}. As a result, pure model-based controller performance is significantly reduced in constrained, unstructured environments. This has motivated an entire family of \textit{model-less} or \textit{data-driven} approaches, such as empirical Jacobian estimation \cite{Yip2014ModelLess}, using Kinematic Model Free approaches for external sensing \cite{ContinuumKMF_Frazelle}, and even disturbance-observer based strategies that compensate for unknown dynamics \cite{DisturbanceObserver2021}. 

In this paper, building forward on the model-less philosophy, our work introduces a novel addition not present in prior literature: \emph{Backbone Tension Optimization} integrated directly into the actuation-level controller. This explicitly regulates the internal backbone axial forces alongside tendon tensions during motion, thereby improving stability, avoiding slack-induced curvature loss, and offering a new control handle not explored in previous model-less frameworks. Combined with online empirical Jacobian adaptation and dual convex updates, our approach provides a robust, fully model-agnostic control scheme capable of precise trajectory tracking without requiring any kinematic or dynamic model of the continuum manipulator.

The paper is organized as follows. Section 2 presents the kinematic modeling of Continuum Manipulators where the formulation of the surrogate forward kinematics and empirical differential kinematics needed to describe the tip motion is described. Section 3 introduces our novel mechanism of Backbone tension optimization for explicitly regulating internal axial tension along with a real-time quadratic program that computes actuator updates for tracking the desired trajectory while enforcing tendon, geometric and tension constraints. Section 3 also presents how the empirical Jacobian is corrected and updated to maintain the differential kinematics. Finally, the simulation results are presented.

\section{Kinematic Modeling of Continuum Manipulators}
Continuum manipulators deform in complex ways due to tendon coupling, internal friction, and contact with unknown obstacles. These effects alter the true task-space Jacobian in ways that cannot be predicted from actuator states alone, causing model-based controllers to become inaccurate or unstable. Therefore, we adopt a simple surrogate geometric model only for initialization, while relying on empirical, data-driven Jacobian updates during operation.

\subsection{Constant Curvature Approximation}
Continuum manipulators are conventionally modeled using a constant curvature assumption, where the backbone  forms an arc of curvature $\kappa$ \cite{ Jones2006Multisection,Webster2010CCReview}. While this model provides simple closed-form kinematics and works well in free space, it fails when the robot interacts with constraints  \cite{Yip2014ModelLess}. Tendon-driven continuum robots typically relate curvature $\kappa$ to differential tendon motion through an approximate affine mapping as shown in Eq. (\eqref{model_cont_manipu}) .
\begin{equation}
x(s) = \frac{1}{\kappa}\big(1 - \cos(\kappa s)\big), \qquad
y(s) = \frac{1}{\kappa}\sin(\kappa s),\label{model_cont_manipu}
\end{equation}
for $s \in [0, L]$ and $L$  is manipulator's Backbone length

In tendon–driven continuum manipulators, the curvature of the manipulator forms due to the differential shortening/pulling of the tendons on each side of the manipulator.  For the simulations in this paper, we have adopted a simplified and effective affine mapping:
\begin{equation}
\kappa = \gamma (y_r - y_\ell),
\end{equation}
where $\gamma$ is the curvature gain. Positive value of $(y_r - y_\ell)$ induces leftward bending, whereas the negative value will induce rightward bending. This captures the bending behavior and is widely used for controller development. Most model–based Jacobians assume a known map $x=f(y)$, however, unmodeled tendon coupling and external forces distort the task-space Jacobian. Since, these effects cannot be inferred from the actuator states alone, analytical Jacobian become difficult to use. This motivates the use of \emph{model–less}, data–driven Jacobian estimation approaches \cite{ContinuumKMF_Frazelle,Yip2014ModelLess, Yip2016ModelLess},  where Jacobian $J$  is updated online from measured motion and they remain robust even under unknown and complex environmental conditions. 

\subsection{Surrogate Forward–Kinematics Model}
\label{sec:surrogate_fk}
A full continuum-robot forward kinematics model requires nonlinear elastic models, tendon geometry and stiffness parameters \cite{Jones2006Multisection, Camarillo2008Mechanics}. As these parameters are difficult to identify and control, we adopt a simple \emph{surrogate} FK map used only for empirical Jacobian initialization. The actuator vector is given by \( y = [\, y_i \; y_\ell \; y_r \,]^\top \)  and the surrogate tip position axial and lateral motion is given by
\begin{equation}
    x = k_x (y_r - y_\ell), \qquad
    y = -k_y\!\left(\frac{y_\ell + y_r}{2} - L\right) + y_i .
    \label{eq:surrogate_fk}
\end{equation}
This model is not used for control but provides consistent tip-displacement predictions for computing initial finite-difference Jacobian columns.

\subsection{Empirical Jacobian Estimation}
Let the end-effector tip cartesian position be denoted by  \( x = f(y) \), where \( f(\cdot) \) represents the unknown forward-kinematic map from actuator commands to the tip motion. Considering a nominal configuration, the differential kinematics is given by \( \dot{x} = J(y)\,\dot{y} \), where \( J(y) \in \mathbb{R}^{2\times 3} \)  represents the instantaneous task-space Jacobian.

To initialize J, each actuator is perturbed independently by a small displacement  $\Delta y_i$ while all others are held fixed. For actuator \( i \), the perturbed configuration is defined as \( y^{(i)} = y + \Delta y_i e_i \).
where $e_i$ is the $i$-th standard basis vector.  
The corresponding tip displacement is measured and is used to construct  Jacobian columns based on the finite‐difference ratio as shown below.
\begin{equation}
    J_{:,i} \approx 
    \frac{\Delta x_i}{\Delta y_i},
    \label{eq:J_fd}
\end{equation}
which matches with the Jacobian update rule proposed in \cite{Yip2014ModelLess, Yip2016ModelLess}. Due to tendon routing and differential gearing, the raw Jacobian often exhibits disproportionate column norms, which can drive ill-conditioning and numerical drift during online inversion. To address this, each column of Jacobian is scaled by its Euclidean norm and collected into a diagonal matrix $W$ as shown in Eq. (\eqref{eq:W_scaling}). 
\begin{equation}
    w_i = \left\|J_{:,i}\right\|_2,
    \qquad 
    W = \mathrm{diag}(w_1, w_2, w_3),
    \label{eq:W_scaling}
\end{equation}
The normalized Jacobian scheme  $\widehat{J} = J W^{-1}$ \cite{Yip2014ModelLess, Yip2016ModelLess}, results in an actuation-balanced differential map that mitigates bias in subsequent optimization.

\section{Backbone Tension Optimization}
Model-less continuum control traditionally enforces only tendon tautness 
\cite{Yip2014ModelLess, Yip2016ModelLess}, ensuring nonzero tension to avoid slack. However, tendon co-activation inevitably induces axial compression along the neutral backbone, which, if unregulated, compromises stiffness, increases curvature hysteresis, and may induce buckling under constrained contact. In this work, internal backbone load is explicitly incorporated into the control objective. 

For a three-tendon system with antagonistic actuation on a single bending axis, the neutral-axis compression is defined as
\begin{equation}
    T_{\mathrm{bb}} = \frac{\tau_\ell + \tau_r}{2},
    \label{eq:Tbb_def}
\end{equation}
where $T_{bb}$, $\tau_\ell$ and $\tau_r$ denote backbone, left- and right-tendon tensions, respectively. Prior methods only constrain the slackness of tendons, our approach explicitly monitors and optimizes internal backbone load $T_{bb}$,  thereby preserving longitudinal stiffness of the continuum manipulator.  Prior model-less implementations enforced $\tau \ge \tau_{min}$  only to avoid slack, whereas explicit computation of  $T_{bb}$  allows real-time regulation of backbone stiffness.
Following the computation of optimal actuation $\Delta y^\star$, tension updates evolve through the linear stiffness map  $\tau^{+} = \tau + K\,\Delta y^\star$ subject to non-slack safety constraint $\tau^{+} \ge \tau_{\min}$. 
The backbone tension is then updated based on updated tendon tensions as shown below. 
\begin{equation}
    T_{\mathrm{bb}}^{+} = \frac{\tau_\ell^{+} + \tau_r^{+}}{2}.
    \label{eq:Tbb_update}
\end{equation}
ensuring that tension minimization is not merely a soft energy objective but a structural consistency condition suppressing compressive saturation during constrained maneuvers. 

\subsection{Optimal Actuation Using Convex Optimization}
\label{sec:optimal_actuation}
Given empirical differential kinematics relating actuator increments to task-space displacement, 
\begin{equation}
    \Delta x \approx J(y)\,\Delta y,
\end{equation}
the controller objective is to compute an actuator update $\Delta y\in\mathbb{R}^3$ that produces a desired Cartesian motion while respecting tendon and geometric limits. The commanded motion is first magnitude-clipped to maintain convergence monotonicity and prevent overshoot.
\begin{equation}
    \Delta x_d = x_d(k) - x(k),
    \label{eq:delta_x_d}
\end{equation}
\begin{equation}
    \Delta x = \min\!\left(1,~\frac{s_{\max}}{\|\Delta x_d\|}\right)\Delta x_d .
    \label{eq:clipped_dx}
\end{equation}
At each cycle, the tendon update $\Delta y^\star$  is obtained from a convex quadratic program \cite{Yip2014ModelLess, Yip2016ModelLess} that simultaneously penalizes task error, tension buildup, and abrupt actuator changes:  
\begin{equation}
\begin{aligned}
\arg\min_{\Delta y}~&
\lambda_x\|J\Delta y - \Delta x\|_2^2
+\lambda_t\|\tau + K\Delta y\|_2^2
+\lambda_y\|\Delta y - \Delta y_{k-1}\|_2^2
\\[2pt]
\text{subject to }~~
&\tau + K\Delta y \ge \tau_{\min},\\
&\Delta y_{\min}\le \Delta y \le \Delta y_{\max},\\
&y_{\min}\le y + \Delta y \le y_{\max},
\end{aligned}
\label{eq:qp}
\end{equation}
where $\lambda_x$ enforces end-effector tracking accuracy, $\lambda_t$ regulates co-activation load consistent with compliant tendon guidance (following \cite{Simaan2008Force}), and $\lambda_y$ suppresses aggressive motor excursions. These inequality constraints simultaneously help maintain tendon tautness, actuator saturation limits, and geometric travel bounds. After solving in real-time using CVX, states propagate as follows:
\begin{equation}
    y(k+1) = y(k) + \Delta y^\star, \qquad
    x(k+1) = f\!\left(y(k+1)\right),
    \label{eq:state_update}
\end{equation}
where $f(\cdot)$ denotes the surrogate forward kinematic map Eq.  \eqref{eq:surrogate_fk}.

\subsection{Online Jacobian Adaptation}
\begin{figure}[H]
\centering
\scalebox{1.2}{
\begin{tikzpicture}[
node distance=1.6cm and 2.0cm,
box/.style={rectangle, rounded corners, draw, thick, align=center,
            minimum width=4.7cm, minimum height=1.2cm},
arrow/.style={-{Latex[length=3mm]}, thick},
]

\node[box, fill=teal!15] (init) {(1) Initialize System\\
$J^0$, $W$};

\node[box, fill=teal!30, below=of init] (qp)
{(2) Optimal Actuation\\
Compute $\Delta y$};

\node[anchor=west, right=2.8cm of qp] (traj)
{\small \shortstack{Reference \\ Trajectory}};

\node[anchor=east, left=4cm of qp] (tendon)
{\small \shortstack{Tendon \\ Sensors}};

\node[box, fill=teal!55, below=of qp] (jac)
{(3) Jacobian Update\\
Compute $\Delta J^\star$};

\node[anchor=west, right=2.8cm of jac] (ee)
{\small \shortstack{End--effector \\ Sensors}};

\draw[arrow] (init) -- node[right]{initial $J^0$, $W$} (qp);


\draw[->] (tendon) -- (qp.west)
    node[midway, above=3pt] {\small commanded $\Delta y^\star$};

\draw[ <-] (qp.east) -- (traj)
    node[midway, above=3pt] {\small desired $\Delta x_d$};

\draw[->] (qp) -- (jac)
    node[midway, right=2pt] {\small applied $\Delta y$};

\draw[->] (ee) -- (jac.east)
    node[midway, above=3pt] {\small measured $\Delta x_{\text{meas}}$};

\draw[arrow] 
    (jac) -- ++(-3,0) |- 
    node[pos=0.8, left=60pt, below=32pt] {\small $J^{k+1}$} 
    (qp);

\end{tikzpicture}
}
\caption{Flowchart of the proposed dual convex-optimization control pipeline with backbone tension regulation.}
\label{fig:flowchart}
\end{figure}
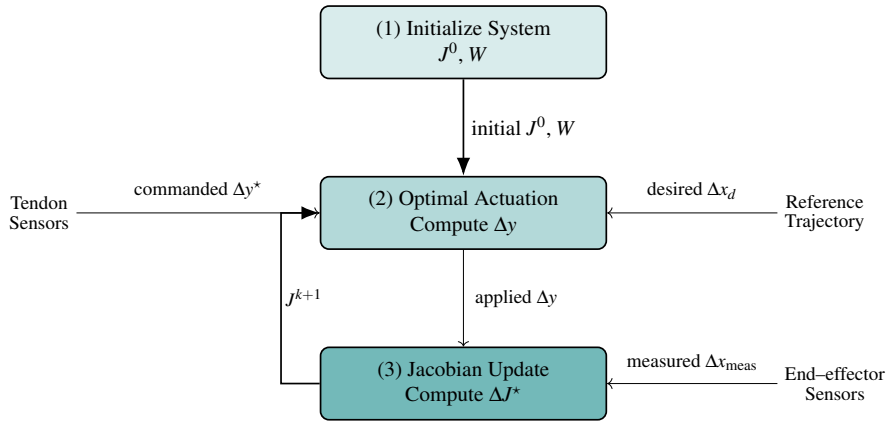

As continuum configuration evolves through backbone bending, insertion, and distributed loading \cite{Jones2006Multisection, Camarillo2008Mechanics}, the Jacobian cannot be assumed static, nor accurately modeled from geometry under contact. Instead, it is estimated online through a minimal-disturbance update at every control step as shown in Fig. ~\ref{fig:flowchart}  ensuring empirical consistency between measured and commanded displacements: 
\begin{equation*}
    \Delta x_{\mathrm{meas}} \approx J(y)\,\Delta y
\end{equation*}

To avoid ill-conditioning, a column-norm matrix $W$ from Eq.  \eqref{eq:W_scaling} is used to map the actuator increments to the normalized space as $ v = W\,\Delta y$.  $W$ is a diagonal column-norm matrix equalizing tendon influence magnitudes. After calculating the Jacobian increment, we minimize the increment $\Delta J$ by solving a convex program
\begin{equation}
\begin{aligned}
\text{minimize}~~
&\|\Delta J\|_F^2 \\
\text{subject to }~& 
\Delta x_{\mathrm{meas}}
= (\widehat{J} + \Delta J)\,v,
\end{aligned}
\label{eq:jac_update}
\end{equation}
To prevent amplification of measurement noise and abrupt kinematic inversions, $\Delta J$ is element-clipped: 
\begin{equation}
    \Delta J_{\mathrm{clip}}
    = \mathrm{sign}(\Delta J^\star)
      \circ \min\!\left(|\Delta J^\star|,\,\Delta J_{\max}\right),
    \label{eq:jac_clip}
\end{equation}
and smoothed via exponential correction: 
\begin{equation}
    \widehat{J}^{+} 
    = \widehat{J} + \alpha_J \Delta J_{\mathrm{clip}}, 
    \qquad
    J^{+} = \widehat{J}^{+} W.
    \label{eq:jac_smooth}
\end{equation}
This ensures $J^{+}$ remains slowly varying even under high constraint curvature, accurately reflecting rotated or attenuated actuation directions when sliding along environmental boundaries, without inducing artificial singularities. 

\section{Results and Discussion}
To validate the performance and robustness of the above proposed dual-convex model-less kinematic controller with online Jacobian adaptation and backbone tension optimization, we conducted a series of simulations using three planar trajectories of increasing complexity: a circle, a regular pentagon and a square. These paths were chosen so as to span smooth curvatures (circle), moderate curvature discontinuity (pentagon) and sharp corners with abrupt changes in direction (square); henceforth testing the controller's ability to handle both continuous and non-smooth geometries. The circular path providing a smooth, continuous trajectory has a reference radius of $80 mm$. The regular pentagon was constructed with a circumradius of $80 mm$ that is, a side length of approximately $94 mm$, thus introducing moderate curvature discontinuities at its edges. Lastly, the square trajectory was defined with a side length of $80 mm$, representing abrupt changes in directions at its four sharp corners. 
\begin{table}[H]
\centering
\caption{Initial conditions and simulation parameters.}
\begin{tabular}{l c}
\hline\hline
\textbf{Parameter} & \textbf{Value} \\
\hline
Backbone length $L$ 
    & $280~\mathrm{mm}$ \\[3pt]

Initial tip position $x(0)$ 
    & $f(y(0)) \approx (0,0)$  \\[3pt]

Slackness bound $\epsilon$ 
    & $0.3~\mathrm{N}$ \\[3pt]

Tendon tension constraints  
    & $0.3~\mathrm{N} \le \tau_i \le 3.0~\mathrm{N}$ \\[3pt]

Stiffness matrix $K$ 
    & $\mathrm{diag}(0.09,\,0.4,\,0.4)$ \\[3pt]

Step-size cap $s_{\max}$ 
    & $1~\mathrm{mm}$ \\[3pt]

Jacobian smoothing gain $\alpha_J$ 
    & $0.15$ \\[3pt]

Jacobian increment clip $\Delta J_{\max}$ 
    & $0.035$ \\[3pt]
\hline\hline
\end{tabular}
\label{tab:initial_conditions}
\end{table}
All experiments start from a straight, unloaded configuration yielding an initial tip position $f(y(0)) = (0,0)$ on the trajectory plane. The initial tendon tensions, stiffness matrix, Jacobian normalization and safety bounds are summarized in Table ~\ref{tab:initial_conditions}. For each of the trajectory, the controller computes optimal actuator increments iteratively through convex optimization while also updating the empirical Jacobian from measured motion, ensuring complete model-less differential kinematics throughout the motion. Further a lower bound of  $0.3\,\mathrm{N}$ in all tendon tensions ($\tau_i \ge 0.3~\mathrm{N}$) is enforced, to prevent slack as it can lead to discontinuous backbone motion and unstable Jacobian updates. Similarly, an upper bound of $3.0\,\mathrm{N}$ ensures that the tension growth remains within the mechanical limits for preserving backbone stiffness.

\begin{table}[H]
\centering
\caption{Parameters of reference trajectory for the three test paths.}
\begin{tabular}{l c}
\hline\hline
\textbf{Trajectory Parameter} & \textbf{Value} \\
\hline
Reference circle radius 
    & $80~\mathrm{mm}$ \\[3pt]

Reference pentagon circumradius 
    & $80~\mathrm{mm}$ \\[3pt]

Reference pentagon side length 
    & $94.0~\mathrm{mm}$ \\[3pt]

Reference square side length 
    & $80~\mathrm{mm}$ \\[3pt]
\hline\hline
\end{tabular}
\label{tab:trajectory_parameters}
\end{table}
\FloatBarrier

\subsection{Tip Trajectory Tracking}

\begin{figure*}[ht]
\centering

\begin{subfigure}[t]{0.32\textwidth}
    \centering
    \includegraphics[width=\linewidth]{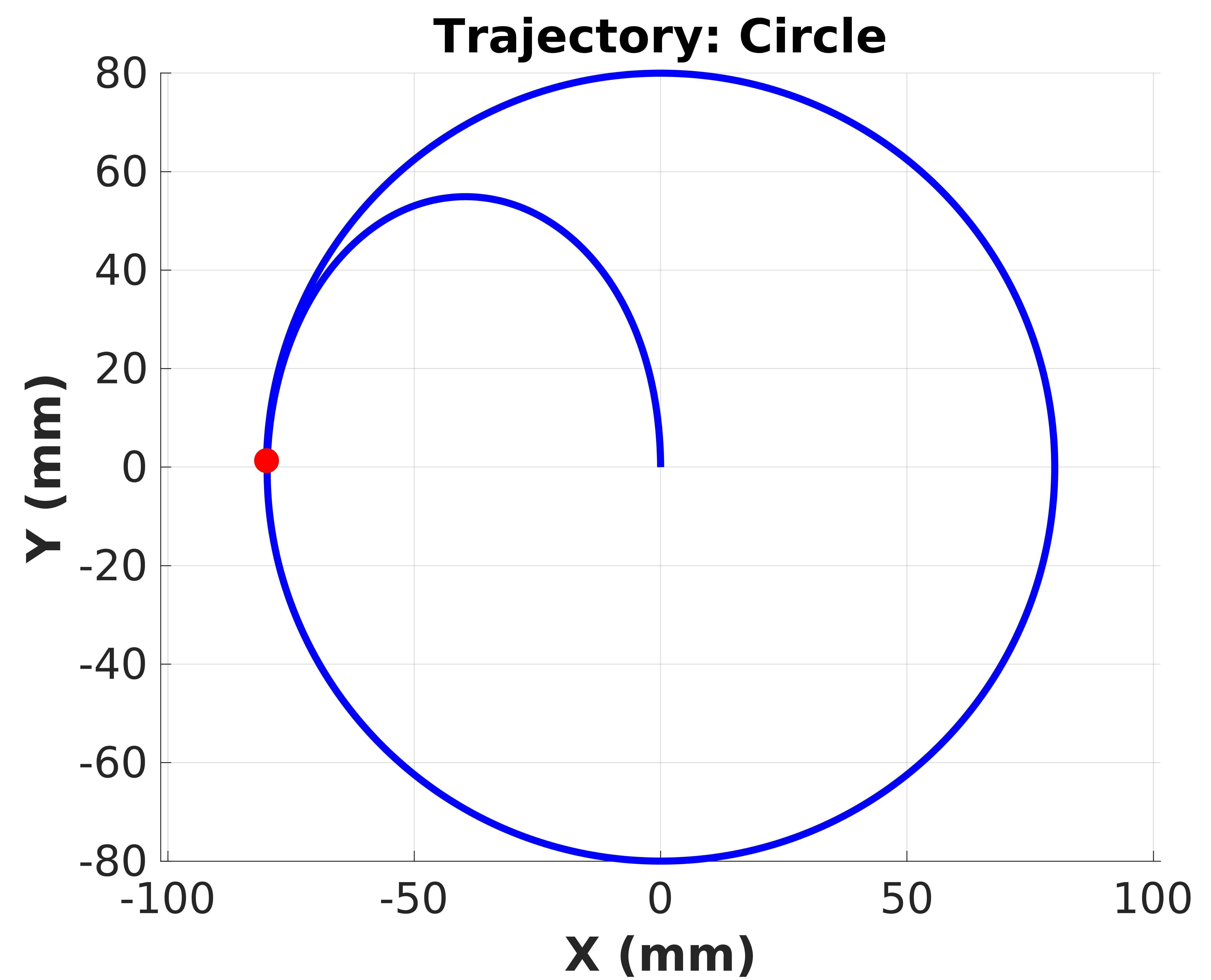}
    \caption{}
    \label{fig:circle_3d_1}
\end{subfigure}
\hfill
\begin{subfigure}[t]{0.32\textwidth}
    \centering
    \includegraphics[width=\linewidth]{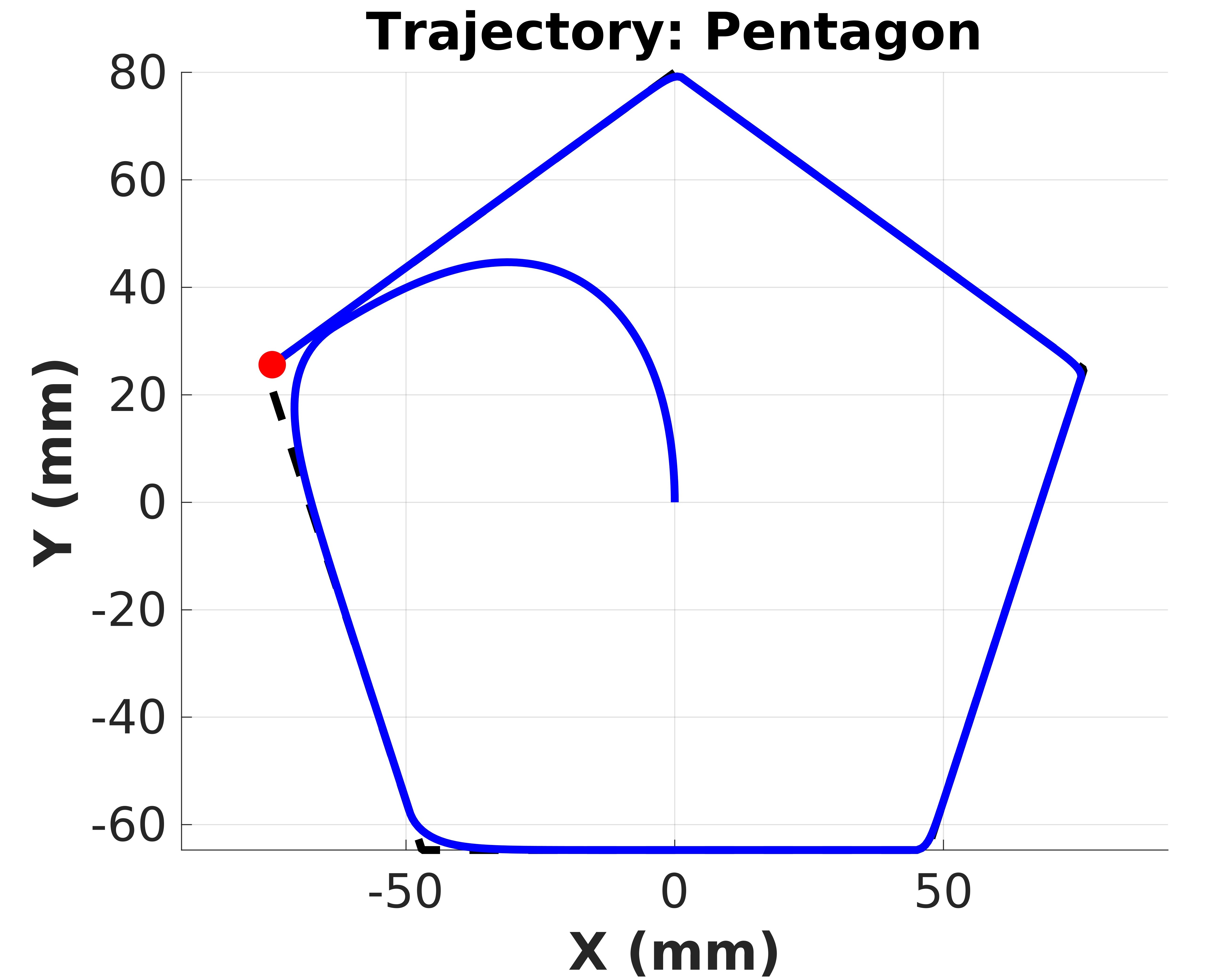}
    \caption{}
    \label{fig:pentagon_3d_1}
\end{subfigure}
\hfill
\begin{subfigure}[t]{0.32\textwidth}
    \centering
    \includegraphics[width=\linewidth]{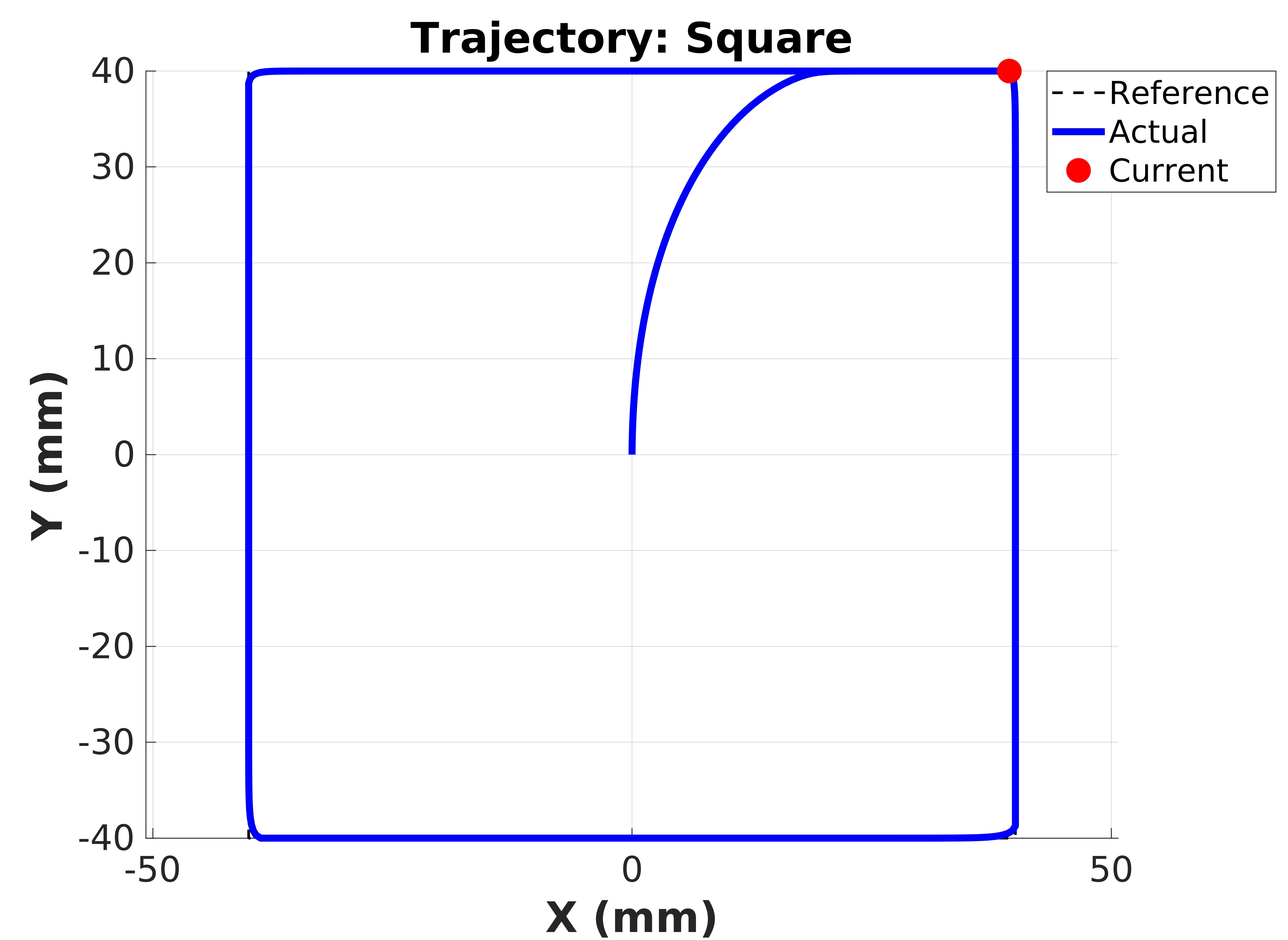}
    \caption{}
    \label{fig:square_3d_1}
\end{subfigure}
\hfill

\caption{Tracked versus reference trajectories for the (a) circle, (b) pentagon, and (c) square paths}
\label{fig:all_3d_snapshots_1}
\end{figure*}

\FloatBarrier
Across all commanded paths, the model-less controller successfully drives the end-effector from its initial straight configuration onto the desired curve, exhibiting smooth convergence and bounded error throughout the motion. Figure 2 reports the tracked versus reference trajectories for circular, pentagonal, and square profiles. In the circular case as shown in Fig. 2(a), the tip converges monotonically toward the reference, with steady-state deviations of approximately $1$-$3$~mm. The helix-like inward convergence indicates that the updated empirical Jacobian rapidly aligns the actuator contributions with the evolving curvature, eliminating overshoot.

For the pentagonal path as shown in Fig. 2(b), the manipulator closely aligns with each linear edge, while the unavoidable finite backbone curvature results in local smoothing at vertices. Nonetheless, after each corner transition, the updated Jacobian stabilizes the tip within $1$--$3$ mm of the corresponding segment. The square trajectory in Fig. 2(c) represents the most demanding curvature transition; while minor disparities occur at discontinuous corners, long straight edges are tracked with high precision, achieving $\approx 0.3$--$0.4$ mm error. These observations collectively demonstrate that, consistent with continuum kinematics, the controller converges toward the lowest-curvature admissible configuration while respecting tendon-driven geometric limits and avoiding artificial singularities.

\subsection{Tendon Tension Evolution}
Figure 3 illustrates the evolution of tendon and backbone tensions. Three behaviors consistently emerge. First, all tendons satisfy the non-slack constraint $\tau_i \ge \epsilon$, preventing loss of controllability in accordance with the slack-avoidance criterion. Second, the system exhibits alternating dominance, i.e., when bending to one side, the corresponding tendon increases proportionally in tension while the antagonist relaxes, yielding smooth bidirectional curvature actuation. Finally, the backbone tension maintains stability through the optimization as established in Section 3. This explicit minimization of co-activation regulates internal compressive loading, suppresses unnecessary tendon force buildup, and avoids buckling under tight curvature transitions. The controller thereby mirrors the tension-balancing effect reported in prior tendon-driven continuum studies, while achieving it \textit{without} reliance on analytical kinematic models. 
\begin{figure*}[ht]
\centering

\begin{subfigure}[t]{0.32\textwidth}
    \centering
    \includegraphics[width=\linewidth]{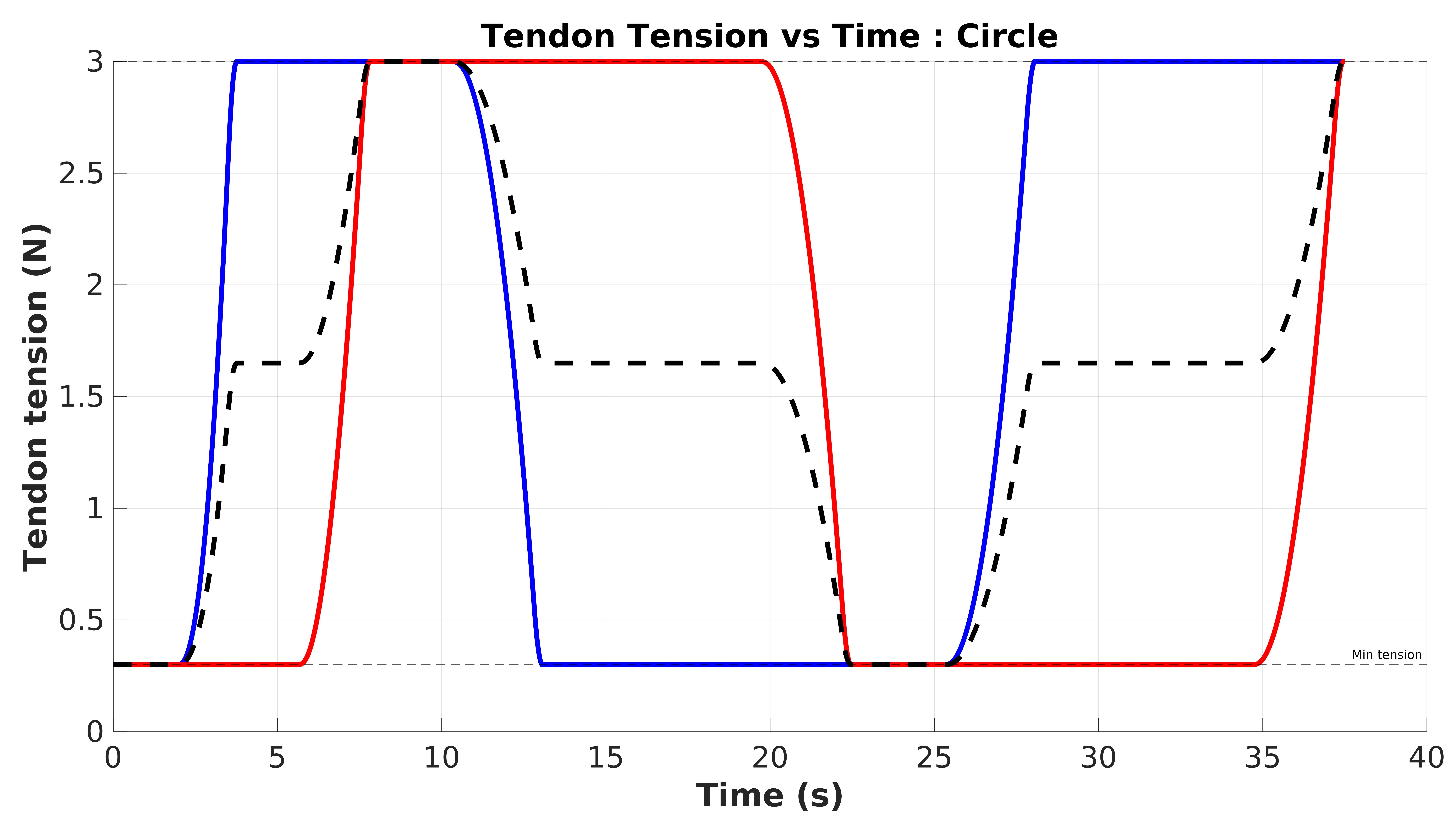}
    \caption{Circle}
    \label{fig:circle_3d_2}
\end{subfigure}
\hfill
\begin{subfigure}[t]{0.32\textwidth}
    \centering
    \includegraphics[width=\linewidth]{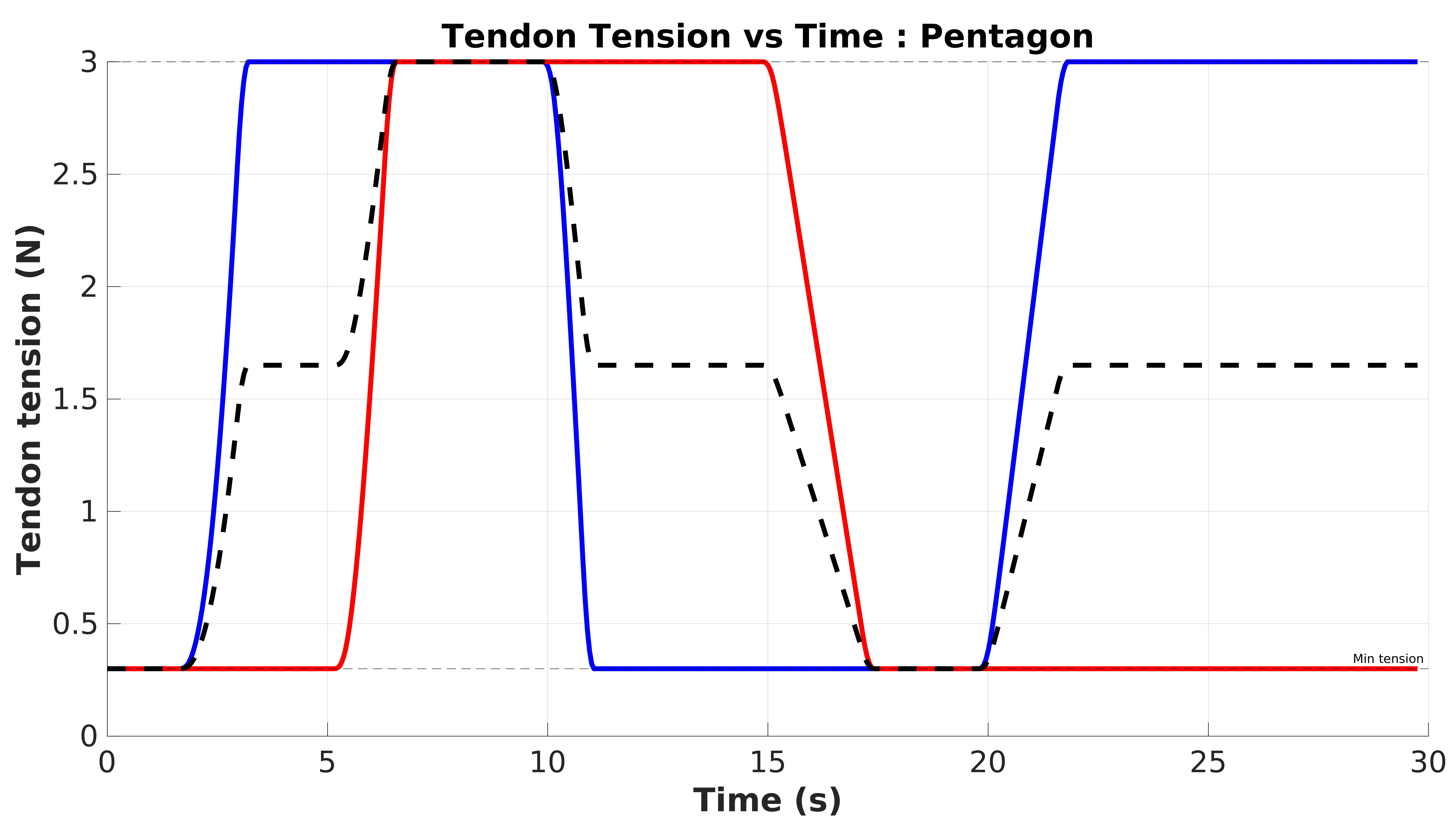}
    \caption{Pentagon}
    \label{fig:pentagon_3d_2}
\end{subfigure}
\hfill
\begin{subfigure}[t]{0.32\textwidth}
    \centering
    \includegraphics[width=\linewidth]{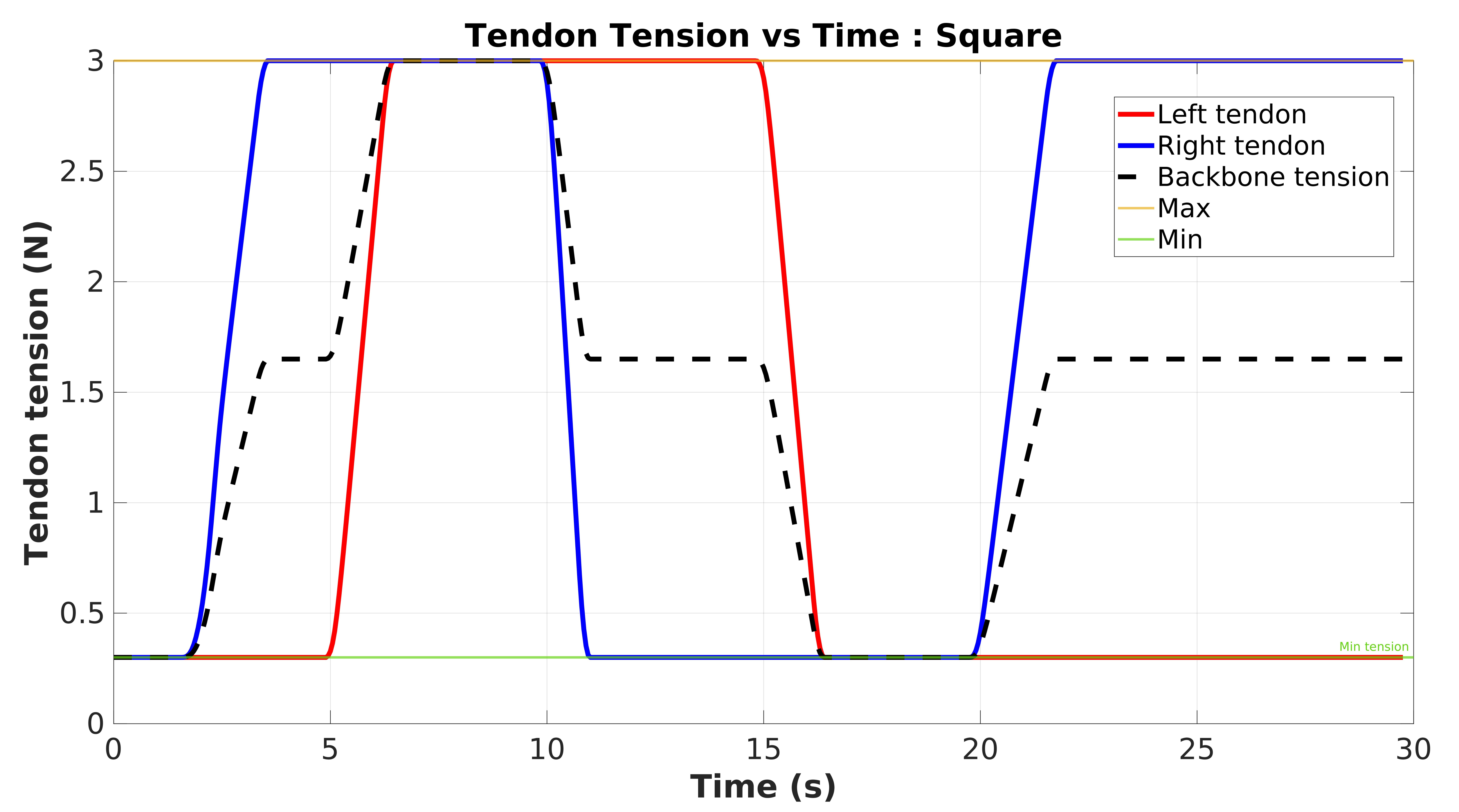}
    \caption{Square}
    \label{fig:square_3d_2}
\end{subfigure}
\hfill

\caption{Tendon–tension evolution for the three reference paths: (a) circle, (b) pentagon, and (c) square paths.}
\label{fig:all_3d_snapshots_2}
\end{figure*}

\FloatBarrier

\subsection{Tracking Error Evolution}
Figure 4 presents the end-effector tracking error as a function of path length traveled along each desired path. The initial error peak is observed in all profiles corresponds to the centric starting pose relative to the reference curve. For the circular trajectory as shown in Fig. 4(a), the error decreases monotonically during convergence and remains nearly constant around the circumference once steady tracking is achieved. In the pentagonal trajectory case, as shown in Fig. 4(b), error minima occur along each linear edge, while localized peaks emerge at vertices due to discrete curvature switching. For the square trajectory as shown in Fig. 4(c), the sharpest curvature discontinuities produce small error spikes at each corner. Once the empirical Jacobian update  $\widehat{J}_{k+1} 
    = \widehat{J} _{k}+ \alpha \Delta J$  converges (with $\alpha$ as the smoothing constant), chattering diminishes and smooth geometric adherence is observed. 

\begin{figure*}[ht]
\centering

\begin{subfigure}[t]{0.32\textwidth}
    \centering
    \includegraphics[width=\linewidth]{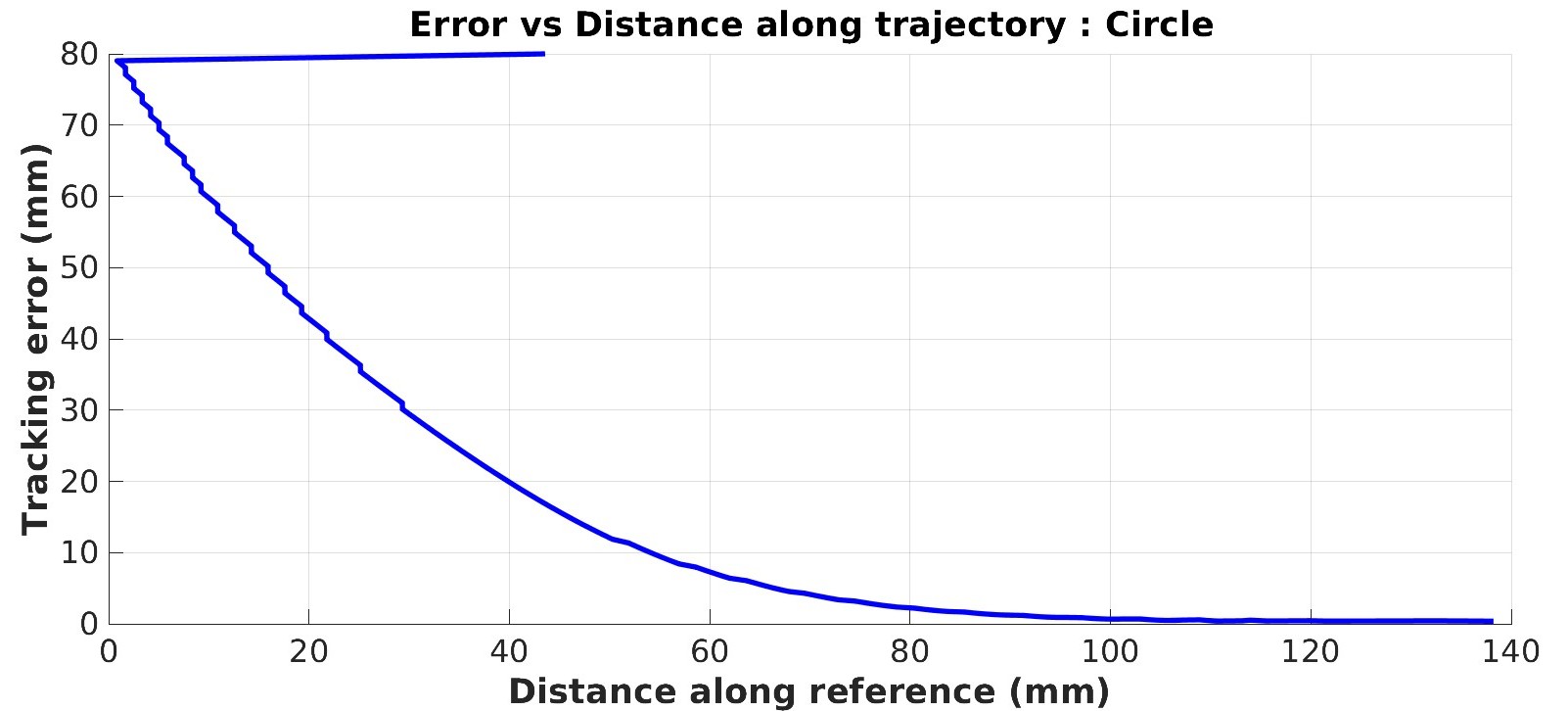}
    \caption{Circle}
    \label{fig:circle_3d_3}
\end{subfigure}
\hfill
\begin{subfigure}[t]{0.32\textwidth}
    \centering
    \includegraphics[width=\linewidth]{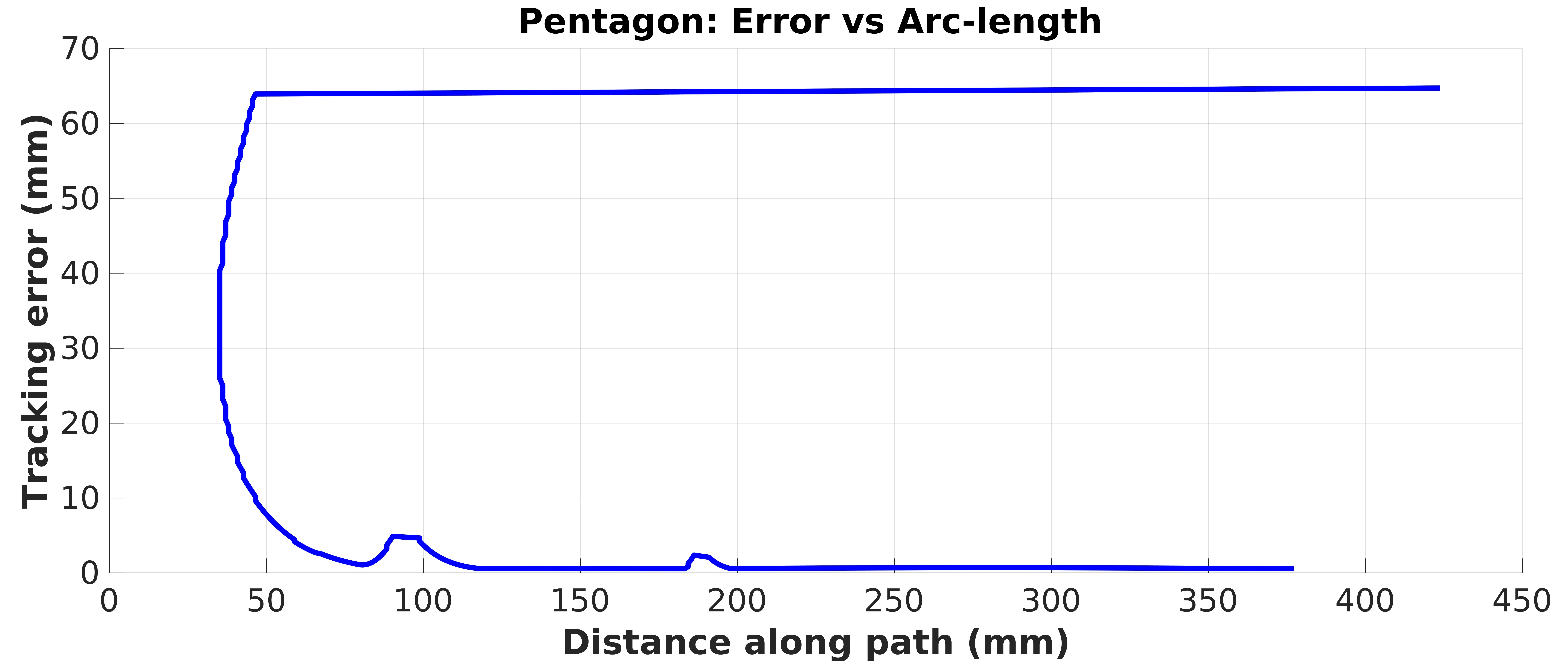}
    \caption{Pentagon}
    \label{fig:pentagon_3d_3}
\end{subfigure}
\hfill
\begin{subfigure}[t]{0.32\textwidth}
    \centering
    \includegraphics[width=\linewidth]{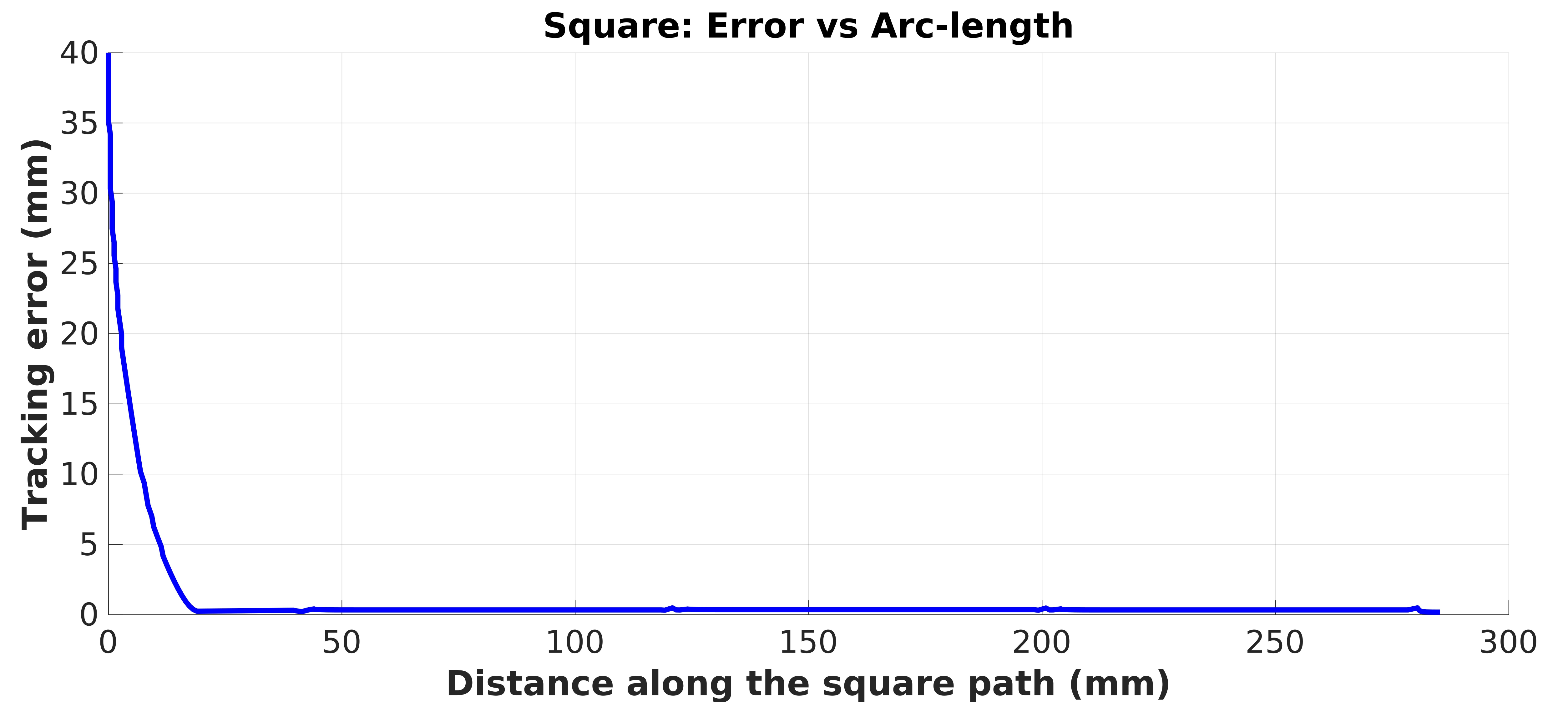}
    \caption{Square}
    \label{fig:square_3d_3}
\end{subfigure}
\hfill

\caption{Error Vs Distance traversed on the trajectories for the (a) circle, (b) pentagon, and (c) square paths}
\label{fig:all_3d_snapshots_3}
\end{figure*}

\FloatBarrier

\subsection{3D Manipulator Snapshots}
To examine spatial backbone deformation, 3D reconstructions are generated with the base fixed and the reference trajectory projected at height of 150 mm. As shown in Fig. 5(a), the circular motion produces a uniform curvature distribution with no localized strain concentrations. In contrast, the pentagonal path of Fig. 5(b) induces curvature localization at each of the five vertices with nearly rectilinear segments between them. The square path in Fig. 5(c) exhibits the most pronounced curvature peaks at the four corners with long straight spans along each edge. These volumetric renderings confirm that the controller generates feasible tendon updates that respect structural bending limits while correctly accommodating discontinuous geometric directives. 

\begin{figure*}[htbp]
\centering

\begin{subfigure}[t]{0.32\textwidth}
    \centering
    \includegraphics[width=\linewidth]{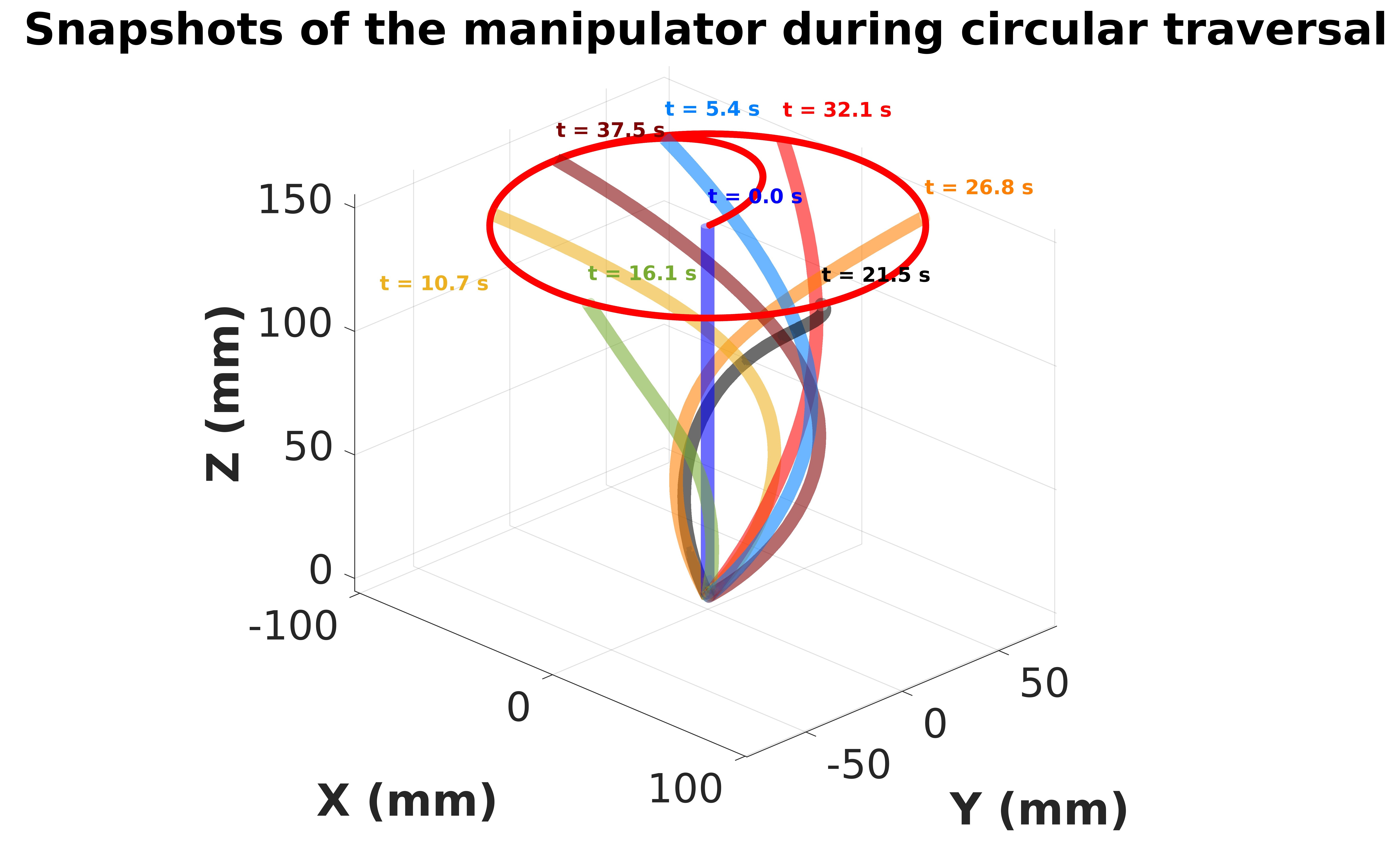}
    \caption{Circle}
    \label{fig:circle_3d_4}
\end{subfigure}
\hfill
\begin{subfigure}[t]{0.32\textwidth}
    \centering
    \includegraphics[width=\linewidth]{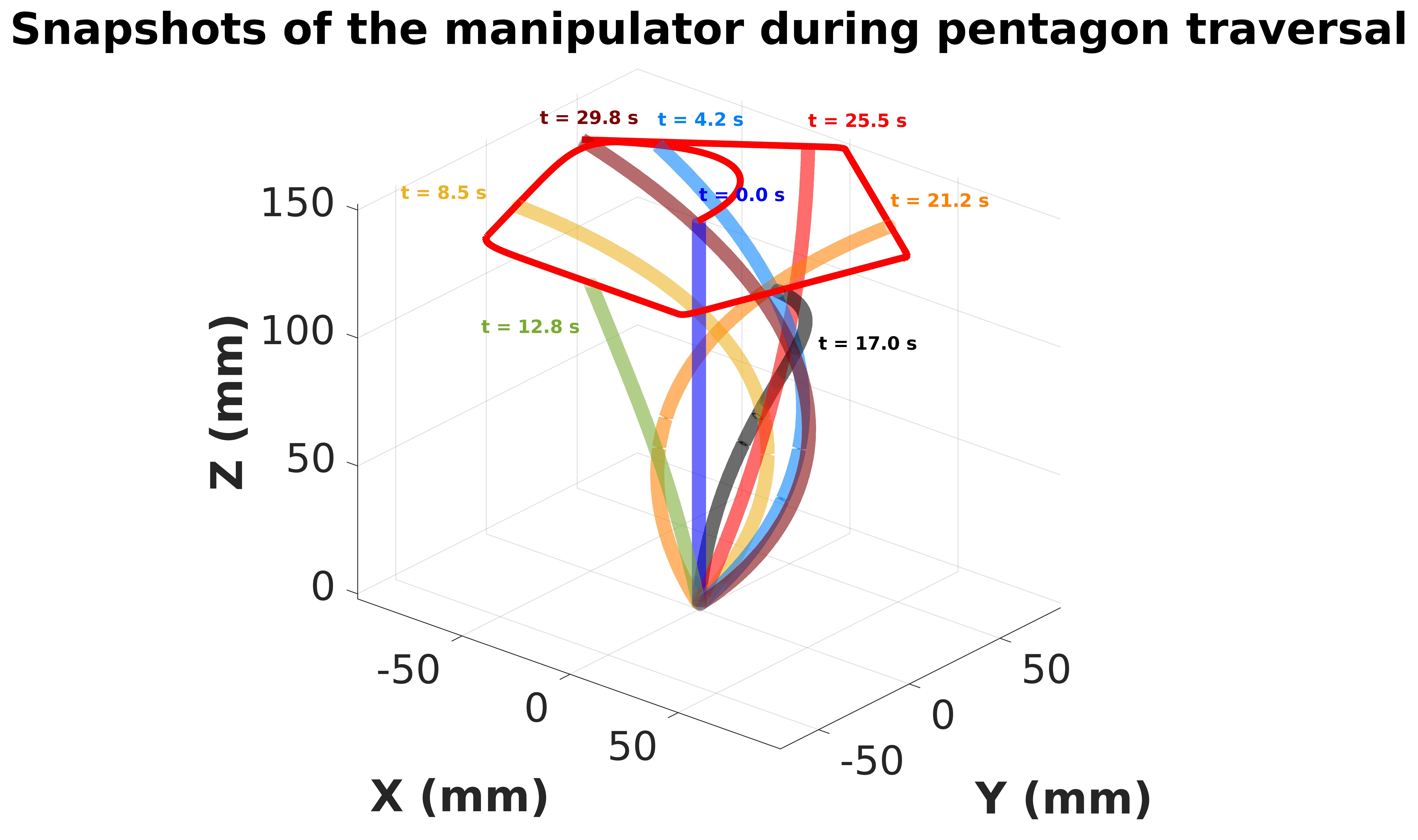}
    \caption{Pentagon}
    \label{fig:pentagon_3d_4}
\end{subfigure}
\hfill
\begin{subfigure}[t]{0.32\textwidth}
    \centering
    \includegraphics[width=\linewidth]{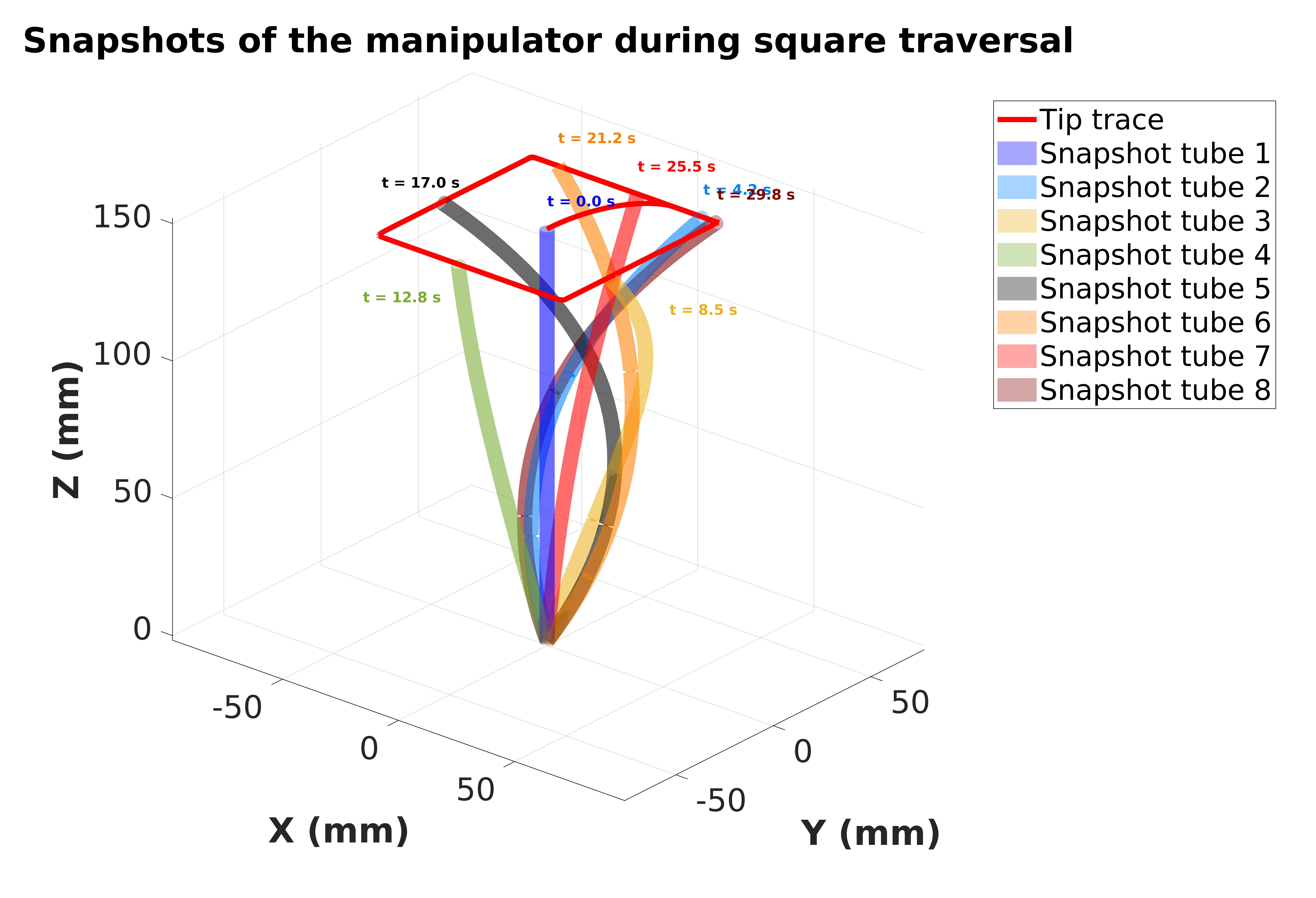}
    \caption{Square}
    \label{fig:square_3d_4}
\end{subfigure}

\caption{3D manipulator snapshots at different timestamps for the three reference trajectories: (a) circle, (b) pentagon, and (c) square.}
\label{fig:all_3d_snapshots_4}
\end{figure*}

\FloatBarrier

\section*{Conclusion}
This study demonstrates a fully model-less continuum control strategy capable of accurately tracking diverse planar trajectories through dual convex optimization, empirical Jacobian adaptation, and active tendon-backbone tension regulation. The incorporation of backbone tension minimization yields additional mechanical robustness, suppressing slack, preventing geometric instability, and maintaining smooth actuation transitions. Future work will extend this framework to full 3D end-effector guidance, incorporate external sensing for occluded environments, and investigate data-driven Jacobian prediction to further reduce estimation lag and improve transient behavior in rapidly changing curvature regimes. 


\end{document}